\documentclass{midl} 

\usepackage{mwe} 
\usepackage{multirow}
\usepackage{adjustbox}

\jmlryear{2025}
\jmlrworkshop{Full Paper -- MIDL 2025}
\jmlrvolume{-- nnn}
\editors{Accepted for publication at MIDL 2025}

\title[PathGenIC]{Histopathology Image Report Generation by Vision Language Model with Multimodal In-Context Learning}

\midlauthor{\Name{Liu, Shih-Wen} \Email{NE6134079@gs.ncku.edu.tw}\\
\addr National Cheng Kung University, Taiwan \AND
\Name{Fan, Hsuan-Yu} \Email{P76124702@gs.ncku.edu.tw}\\
\addr National Cheng Kung University, Taiwan \AND
\Name{Chu, Wei-Ta} \orcid{0000-0001-5722-7239} \Email{wtchu@gs.ncku.edu.tw}\\
\addr National Cheng Kung University, Taiwan \AND
\Name{Yang, Fu-En} \Email{fredy@nvidia.com}\\
\addr NVIDIA Research \AND
\Name{Wang, Yu-Chiang Frank} \Email{frankwang@nvidia.com}\\
\addr NVIDIA Research \\
}

\begin{document}

\maketitle

\begin{abstract}
Automating medical report generation from histopathology images is a critical challenge requiring effective visual representations and domain-specific knowledge. Inspired by the common practices of human experts, we propose an in-context learning framework called PathGenIC that integrates context derived from the training set with a multimodal in-context learning (ICL) mechanism. Our method dynamically retrieves semantically similar whole slide image (WSI)-report pairs and incorporates adaptive feedback to enhance contextual relevance and generation quality. Evaluated on the HistGen benchmark, the framework achieves state-of-the-art results, with significant improvements across BLEU, METEOR, and ROUGE-L metrics, and demonstrates robustness across diverse report lengths and disease categories. By maximizing training data utility and bridging vision and language with ICL, our work offers a solution for AI-driven histopathology reporting, setting a strong foundation for future advancements in multimodal clinical applications. 
\end{abstract}

\begin{keywords}
Multimodal In-Context Learning, Medical Report Generation, Histopathology Images, Vision-Language Models, HistGen Benchmark.
\end{keywords}

\section{Introduction}
\label{sec:introduction}
Histopathology reports are important diagnostic tools, providing detailed interpretations of histopathological findings that directly influence patient care. Composing a histopathology report demands significant expertise and is very time-consuming. Therefore, automating this process holds immense potential to enhance diagnostic efficiency and reduce workloads. Yet, this task is far from trivial, as it requires understanding the intricate visual features of histopathology images and generating accurate and structured documents.

Recent advancements in medical AI have significantly improved the integration of visual and textual data for histopathology report generation. Models like HistGen \citep{histgen} and WsiCaption \citep{wsicaption} have made strides in bridging the gap between whole slide images (WSIs) and histopathology reports through multimodal approaches. Similarly, models like Quilt-LLaVA \citep{quilt_llava} and LLaVA-Med \citep{llavamed} integrate medical imaging and language models for broader applications, including Visual Question Answering (VQA) and histopathology report generation. Despite these advances, these medical models overlook utilizing the similarity between a test image and the images (and their associated histopathology reports) in the training datasets. 

Taking similar images and their associated reports from previously confirmed cases as the reference is a common practice for expert pathologists to make a report efficiently and ensure report quality. Reviewing similar cases can facilitate maintaining diagnostic accuracy and consistency \citep{pathology_peer_review}. They also rely on the Diagnosis Learning Cycle \citep{diagnosis_learning_cycle}, a conceptual framework designed to enhance diagnostic performance through feedback, reflection, and continuous learning, emphasizing the importance of learning from past experiences. Inspired by this, we propose to retrieve similar images and their associated reports from the database and then take them as important clues for generating a report of the test image. We employ a fine-tuned histopathology-specific vision language model (VLM) and investigate prompting techniques to generate accurate histopathology reports. 

We evaluate our framework on the HistGen benchmark, achieving state-of-the-art performance across BLEU, METEOR, and ROUGE-L metrics. We also test the proposed components separately and study performance obtained based on different settings. In summary, our contributions include:
\begin{itemize}
    \item Introducing an expert-inspired framework with multimodal in-context learning to utilize information from the training dataset to generate better histopathology reports.
    \item Fine-tuning a histopathology-specific VLM to enhance its capabilities in generating histopathology reports.
    \item Achieving state-of-the-art results on the HistGen benchmark and demonstrating the robustness and efficacy across various configurations. 
\end{itemize}


\section{Related Works}
\paragraph{Medical Report Generation.}
Medical report generation has received significant attention recently. Models like HistGen \citep{histgen} and WsiCaption \citep{wsicaption} have demonstrated success in generating histopathology reports by bridging WSIs and textual data through advanced encoding and generative approaches. HistGen employs local-global feature encoding to enhance contextual coherence in reports. WsiCaption uses a multiple-instance generative model to produce detailed clinical descriptions. Vision-language models like Quilt-LLaVA \citep{quilt_llava} and LLaVA-Med \citep{llavamed} extend these capabilities to broader biomedical applications. LLaVA-Med adapts the LLaVA framework to handle diverse medical tasks, including report generation and VQA, by leveraging curriculum learning to align visual and textual data. Quilt-LLaVA focuses on histopathology-specific applications, aligning representations of WSIs and text using histopathology-specific datasets like Quilt-Instruct. While these models achieve automatic report generation, they often overlook contextual information from training datasets, diverse histopathology categories, or the common practices of expert pathologists.

\paragraph{In-Context Learning and Retrieval-Augmented Generation.}
In-context learning (ICL) has emerged as a promising technique for improving model adaptability without retraining. Pioneered by models like GPT-3 \citep{gpt3}, ICL enables models to work on tasks by incorporating task-specific examples directly into the input prompts. Retrieval-augmented generation (RAG) \citep{rag} is one of the ICL strategies that embodies this concept by utilizing retrieved data to increase model adaptability. While ICL and RAG have shown success in general-purpose language models, their application to medical AI remains limited. Existing multimodal models like Quilt-LLaVA and LLaVA-Med focus on directly encoding visual and textual data but lack robust mechanisms for retrieving and leveraging similar examples for better generation. Our work attempts to bridge the gaps by incorporating ICL into a report generation framework based on context derived from similar training examples, category-specific guidelines, and structured feedback.


\section{Methodology}
\subsection{Base Model}
\label{sec:base}
\begin{figure}
    \centering
    \includegraphics[width=13cm]{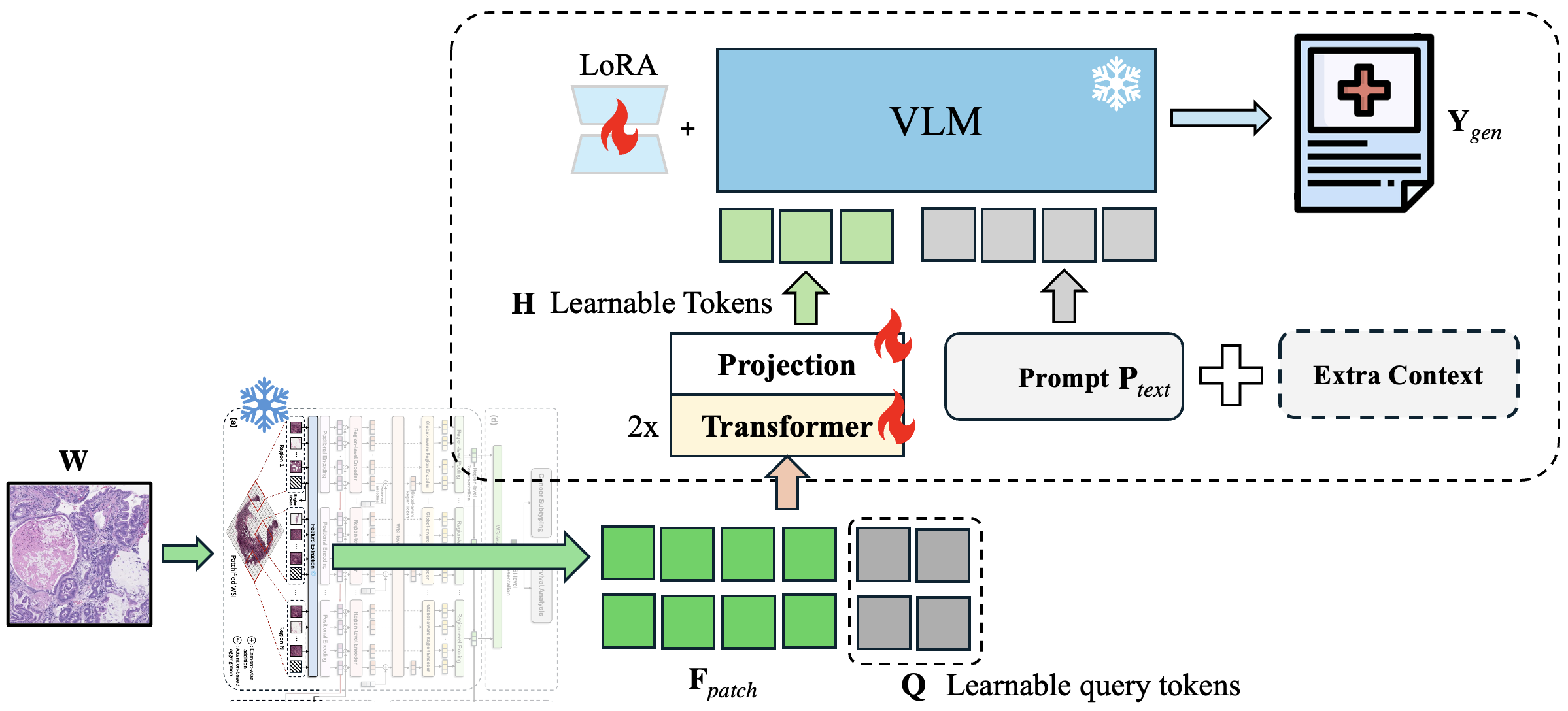}
    \caption{Overview of the proposed framework. WSI patch features $\mathbf{F}_{patch}$ and the learnable tokens $\mathbf{Q}$ are jointly processed by a transformer to get holistic WSI features $\mathbf{H}$. The processed $\mathbf{H}$ with the text prompts $\mathbf{P}_{text}$ are then fed to a VLM to generate a report $\mathbf{Y}_{gen}$. Our main contribution is enriching prompts with extra context. The components with flame symbols mean that we need to train or fine-tune the parameters, and the ones with snowflake symbols mean that we adopt the parameters pre-trained by existing works and these parameters are frozen in the training process. }
    \label{fig:framework}
\end{figure}

Figure~\ref{fig:framework} illustrates the proposed report generation framework. We take Quilt-LLaVA as the base model, which has been trained on diverse histopathology datasets for VQA tasks. It was trained based on histopathology image patches and was dedicated to VQA. To more effectively extract features from the entire histopathology image and focus on report generation, we replace its visual encoder with Vision Transformer Large (ViT-L) provided in HistGen \citep{histgen}. The ViT-L has been extensively pre-trained using the DINOv2 technique and is used to extract WSI\footnote{All histopathology data are represented in whole slide images in this work. Therefore, we use histopathology image and whole slide image interchangeably.} features $\mathbf{F}_{patch} \in \mathbb{R}^{n \times d}$, where $n$ is the number of patches in a WSI and $d$ is the feature dimension. 

The extracted patch-based features $\mathbf{F}_{patch}$ and $m$ learnable query tokens $\mathbf{Q} \in \mathbb{R}^{m \times d}$ are processed by transformer blocks. The goal is to consider the context between patches to form holistic WSI features: 
\begin{equation}
\hat{\mathbf{H}} = \text{TransformerBlock}(\mathbf{Q}, \mathbf{F}_{patch}),
\end{equation}
where $\hat{\mathbf{H}} \in \mathbb{R}^{m \times d}$ represents the $m$ contextualized tokens derived from $\mathbf{Q}$. Referring to Figure~\ref{fig:framework}, the contextualized tokens $\hat{\mathbf{H}}$ are processed by a projector to be the embeddings more appropriate to a VLM, and the results are denoted as $\mathbf{H} \in \mathbb{R}^{m \times d}$ in the following. Notice that the number of patches $n$ in different WSIs may differ. But we always represent a WSI by the $m$ contextualized tokens $\mathbf{H}$. 

The tokens $\mathbf{H}$ are then passed to a VLM along with a text prompt $\mathbf{P}_{text}$ to generate a histopathology report $\mathbf{Y}_{gen}$:
\begin{equation}
\mathbf{Y}_{gen} = \text{VLM}(\mathbf{H}, \mathbf{P}_{text}). 
\end{equation}
The text prompt $\mathbf{P}_{text}$ (shorten version) for the baseline model is: \texttt{What are the diagnostic findings in the image? Provide a professional, accurate, and well-structured histopathology report for it.}

\subsection{In-Context Learning}
\label{sec:incontext}
\begin{figure}
    \centering
    \includegraphics[width=15cm]{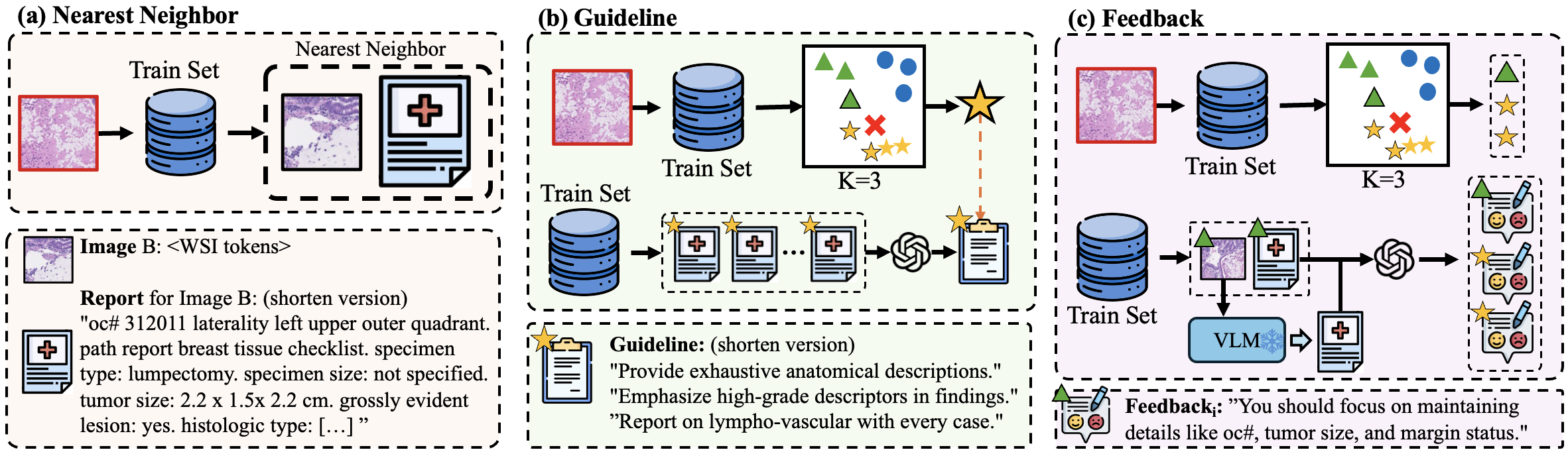}
    \caption{Illustrations of three different clues for in-context learning.}
    \label{fig:concept}
\end{figure}

Based on the base model described above and the framework shown in Figure~\ref{fig:framework}, we propose to extract extra context from the training dataset to facilitate better report generation. The three in-context learning mechanisms are nearest neighbor, category guideline, and feedback. Figure~\ref{fig:concept} illustrates these three different mechanisms. 

\paragraph{Nearest Neighbor.}
In this mechanism, we retrieve the WSI-report pair that is closest to the test WSI from the training set. We calculate the cosine similarity between the WSI tokens of the test image $\mathbf{H}_{test}$ and every WSI in the training set $\mathbf{H}_{train}$.
The image tokens of the WSI most similar to $\mathbf{H}_{test}$, denoted as $\mathbf{H}_{ret}$, are appended to the test WSI tokens, i.e., $ \mathbf{H}_{nn} = (\mathbf{H}_{test}, \mathbf{H}_{ret})$. The histopathology report $\mathbf{Y}_{ret}$ corresponding to $\mathbf{H}_{ret}$ is appended to the text prompt, i.e., $\mathbf{P}_{nn} = (\mathbf{P}_{text}, \mathbf{Y}_{ret})$. 

Given $\mathbf{H}_{nn}$ and $\mathbf{P}_{nn}$, we guide the VLM to generate better histopathology reports with the visually similar WSI tokens and their corresponding report as a reference, i.e., 
\begin{equation}
\mathbf{Y}_{gen} = \text{VLM}(\mathbf{H}_{nn}, \mathbf{P}_{nn}). 
\label{eq:nn} 
\end{equation}

\paragraph{Category Guideline.}
The WSIs in the HistGen benchmark are classified into 32 categories according to disease types. Each WSI is labeled with a disease type. Considering cues from the disease category that is closest to the test WSI may benefit report generation. To do this, we compare the test WSI tokens $\mathbf{H}_{test}$ with WSIs in the training set. We check the disease types of the $K$ WSIs that are most similar to $\mathbf{H}_{test}$. The major category of these $K$ WSIs is then determined. Let's denote it as the category $C$.   

There are multiple WSIs and their associated reports in the category $C$. We then ask GPT-4o\footnote{The exact version is gpt-4o-2024-08-06.} to review these medical reports and generate a representative guideline $\mathbf{T}_{rep}$. With this clue, we finally ask a VLM to generate a report of the test WSI by giving it $\mathbf{H}_{test}$, $\mathbf{T}_{rep}$, and $\mathbf{P}_{text}$, i.e.,   
\begin{equation}
\mathbf{Y}_{gen} = \text{VLM}(\mathbf{H}_{test}, \mathbf{T}_{rep}, \mathbf{P}_{text}). 
\label{eq:guideline} 
\end{equation}

\paragraph{Feedback.} 
Generated reports are usually not perfect. Considering the difference between a generated report and its perfect counterpart may provide clues to enhance report generation. Specifically, we first generate a report for each WSI in the training set by the base model. For each WSI, we then ask GPT-4o to compare the generated report with the corresponding truth report and generate feedback about how to improve the generated version. For each WSI $W_i$, we thus have the corresponding feedback $\mathbf{B}_i$. We believe that report generation can benefit from feedback, as emphasized in the Diagnosis Learning Cycle. 

During testing, we find the $K$ WSIs most similar to the test WSI and obtain the $K$ feedback. A VLM is then asked to generate a report for the test WSI by giving the $K$ feedback and $K$ retrieved WSI tokens, i.e., 
\begin{equation}
\mathbf{Y}_{gen} = \text{VLM}(\mathbf{H}_{test}, \{\mathbf{B}_j\}_{j=1}^K, \mathbf{P}_{text}). 
\label{eq:feedback} 
\end{equation}

The aforementioned three contexts, including nearest neighbor, category guideline, and feedback, can be used for in-context learning separately. We can also jointly use them by inputting all elements mentioned in eqn.~(\ref{eq:nn}), eqn.~(\ref{eq:guideline}), and eqn.~(\ref{eq:feedback}). In the following, the full version of the proposed model is called \textbf{PathGenIC}, standing for HistoPathology report Generation with In-Context learning.  

\subsection{Loss Function and Training}
The model is trained based on the cross-entropy loss to align the generated report $\mathbf{Y}_{gen}$ with the ground truth report $\mathbf{Y}_{true}$. During training, the components with the flame symbols in Figure~\ref{fig:framework} are trained to complete this framework. The fine-tuned parts include the transformer blocks to process WSI patch tokens and learnable query tokens and the LoRA \citep{lora} component that adapts the VLM, i.e., Quilt-LLaVA, to generate reports.



\section{Experiments}
\subsection{Implementation Details}
\paragraph{Benchmark: HistGen Dataset.}
We use the HistGen dataset \citep{histgen} for evaluating report generation. It contains 7,690 WSIs and their corresponding histopathology reports collected from TCGA, spanning 32 disease categories. It is divided into training (80\%), validation (10\%), and test (10\%) subsets, resulting in 6,152 training samples, 769 validation samples, and 769 test samples. 

To fairly compare with existing methods on the HistGen dataset, we adopt BLEU, METEOR, and ROUGE-L as the evaluation metrics. These metrics collectively measure lexical similarity, semantic relevance, and structural coherence between the generated and ground-truth reports. However, they were proposed from the natural language processing perspective and may not well reflect domain entities or inferential consistency \cite{miura21}. To enhance the evaluation, we further show performance in terms of Exact Entity Match Reward ($\text{fact}_{\text{ENT}}$) proposed in \cite{miura21}, which captures the completeness of a generated report by measuring its coverage of entities. 

\paragraph{Experiments Setup.}
We use a batch size of 8 and train the model for 20 epochs. The Adam optimizer is employed with an initial learning rate of $1 \times 10^{-4}$, which follows a cosine reduction schedule to zero. These settings are designed empirically to ensure performance convergence.  Quilt-LLaVA is selected as the base model because it is pre-trained on a big histopathology image dataset for the VQA task. This provides a good foundation for us to extend to histopathology report generation. All experiments are conducted on 2 NVIDIA RTX3090 GPU with 24 GB memory. The implementation is based on PyTorch. To calculate the value of $\text{fact}_{\text{ENT}}$, we adopt BioBERT-v1.1 as the named entity recognition model. 

\begin{table}
\centering
\caption{Performance comparison of different methods on the HistGen benchmark.}
\begin{adjustbox}{max width=\textwidth}
\begin{tabular}{l|lcccccc | c}
\hline
\textbf{Feature Extractor} & \textbf{Methods} & \textbf{BLEU-1} & \textbf{BLEU-2} & \textbf{BLEU-3} & \textbf{BLEU-4} & \textbf{METEOR} & \textbf{ROUGE-L} & $\text{fact}_{\text{ENT}}$ \\
\hline
\hline
\multirow{6}{*}{ResNet50} 
& \textbf{Show\&Tell} \citep{showtell} & 0.249 & 0.099 & 0.047 & 0.025 & 0.086 & 0.165 &\\
& \textbf{UpDownAttn} \citep{updown} & 0.250 & 0.115 & 0.065 & 0.043 & 0.096 & 0.180 &\\
& \textbf{Transformer} \citep{vaswani17}& 0.249 & 0.114 & 0.065 & 0.042 & 0.095 & 0.176 &\\
& \textbf{M2Transformer} \citep{r2trans} & 0.250 & 0.115 & 0.065 & 0.042 & 0.095 & 0.180 &\\
& \textbf{R2Gen} \citep{r2gen} & 0.240 & 0.105 & 0.058 & 0.036 & 0.089 & 0.177 &\\
& \textbf{R2GenCMN} \citep{r2gencmn} & 0.225 & 0.095 & 0.047 & 0.022 & 0.094 & 0.151 & \\
\hline
\multirow{6}{*}{CTransPath} 
& \textbf{Show\&Tell} \citep{showtell}& 0.262 & 0.126 & 0.071 & 0.043 & 0.094 & 0.184 &\\
& \textbf{UpDownAttn} \citep{updown}& 0.240 & 0.139 & 0.090 & 0.063 & 0.100 & 0.201 &\\
& \textbf{Transformer} \citep{vaswani17}& 0.271 & 0.165 & 0.112 & 0.082 & 0.113 & 0.227 &\\
& \textbf{M2Transformer} \citep{r2trans}& 0.259 & 0.160 & 0.108 & 0.076 & 0.103 & 0.218 &\\
& \textbf{R2Gen} \citep{r2gen}& 0.237 & 0.135 & 0.085 & 0.054 & 0.086 & 0.205 &\\
& \textbf{R2GenCMN} \citep{r2gencmn}& 0.211 & 0.098 & 0.054 & 0.033 & 0.079 & 0.158 & \\
\hline
\multirow{7}{*}{DINOv2 ViT-L} 
& \textbf{Show\&Tell} \citep{showtell}& 0.189 & 0.094 & 0.056 & 0.039 & 0.070 & 0.165 &\\
& \textbf{UpDownAttn} \citep{updown}& 0.320 & 0.206 & 0.147 & 0.112 & 0.131 & 0.271 &\\
& \textbf{Transformer} \citep{vaswani17}& 0.382 & 0.266 & 0.200 & 0.157 & 0.162 & 0.316 &\\
& \textbf{M2Transformer} \citep{r2trans}& 0.321 & 0.213 & 0.152 & 0.112 & 0.131 & 0.266 &\\
& \textbf{R2Gen} \citep{r2gen}& 0.274 & 0.166 & 0.107 & 0.071 & 0.102 & 0.234 & \\
& \textbf{HistGen} \citep{histgen} & 0.413 & 0.297 & 0.229 & 0.184 & 0.182 & 0.344 & \\
\hline
\multirow{2}{*}{DINOv2 ViT-L}
& \textbf{Base (Ours)} & \textbf{0.411} & \textbf{0.290} & \textbf{0.222} & \textbf{0.178} & \textbf{0.184} & \textbf{0.336} & 0.445 \\
& \textbf{PathGenIC (Ours)} & \textbf{0.431} & \textbf{0.313} & \textbf{0.243} & \textbf{0.196} & \textbf{0.197} & \textbf{0.357} & 0.462\\
\hline
\end{tabular}
\end{adjustbox}
\label{tab:main_results}
\end{table}

\subsection{Performance of Report Generation}
Table~\ref{tab:main_results} shows a performance comparison between different methods on the HistGen benchmark. As can be seen, both our base model and the PathGenIC model outperform existing methods. With the in-context learning mechanism, the PathGenIC method improves the base model, showing the benefits brought by contextual clues. We also show the values of $\text{fact}_{\text{ENT}}$ obtained by our methods in the rightmost column, while the values of other methods are not available.  

\begin{figure}
\centering
\includegraphics[width=7cm]{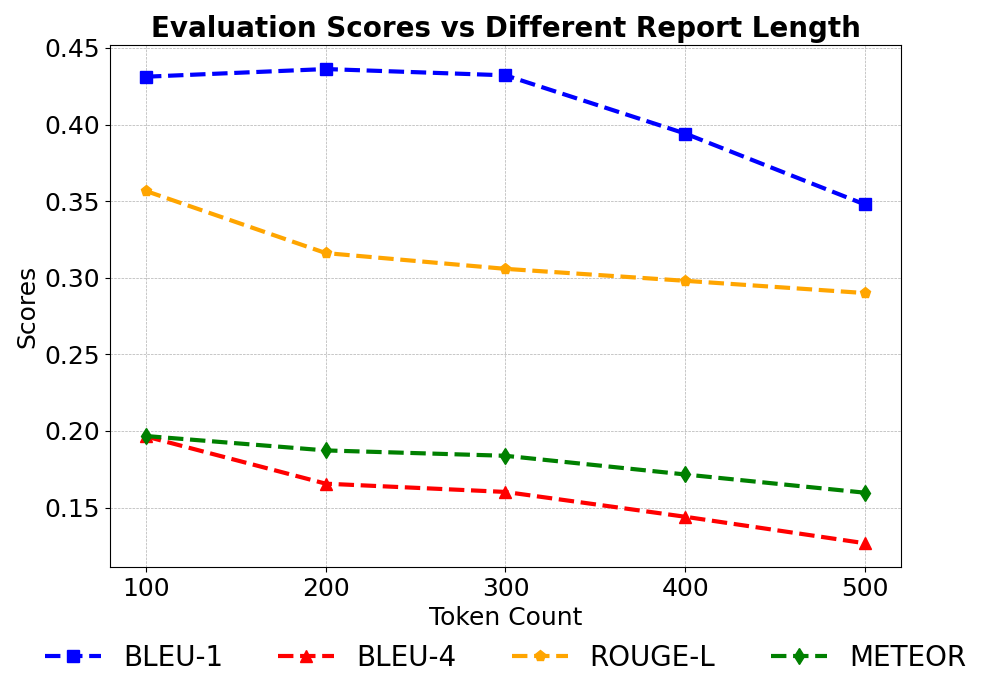}
\caption{Performance across varying sequence lengths (100 to 500 tokens).}
\label{fig:bleu_vs_tokens}
\end{figure}

\paragraph{Analysis of Report Length.}
The results of all methods shown in Table~\ref{tab:main_results} are obtained based on only the first 100 tokens of the generated reports, roughly 80 to 90 words. To understand the proposed PathGenIC more deeply, we study how the report's length influences performance. Figure~\ref{fig:bleu_vs_tokens} shows that most metrics slightly decline as the report's length increases, but the overall quality remains high. Interestingly, we observe a slight improvement in the BLEU-1 scores as the report length increases from 100 to 200 tokens. This suggests that slightly longer reports may enhance the likelihood of word matches between the generated and ground truth reports, possibly due to a more extensive context allowing for better word choice predictions. As the report length increases more, the occurrence of less common words rises and makes a downward trend in metrics. 

\paragraph{Ablation Study of \#Retrieved WSIs.}
We retrieve the top $K$ WSIs and their associated reports as the reference in the category guideline and the feedback mechanisms. We evaluate how different $K$'s influence performance in Table~\ref{tab:abl_k}. The results show that using $K = 3$ gives the best performance. Using $K = 1$ offers comparable performance with slightly reduced effectiveness. This reduction can be attributed to the instability of clues when fewer cases are considered. Conversely, using $K = 5$ introduces noise into the system and negatively affects the model's performance.

\begin{table}
\centering
\footnotesize
\caption{Performance variations when different numbers of nearest neighbors ($K$) in WSI-report pair retrieval.}
\begin{tabular}{lcccc}
\hline
\textbf{Method} & \textbf{BLEU-1} & \textbf{BLEU-4} & \textbf{METEOR} & \textbf{ROUGE-L} \\
\hline
$K$=1 & 0.431 & 0.195 & 0.196 & 0.356 \\
$K$=3 & 0.431 & 0.196 & 0.197 & 0.357 \\
$K$=5 & 0.428 & 0.192 & 0.195 & 0.352 \\
\hline
\end{tabular}
\label{tab:abl_k}
\end{table}

\begin{table}
\centering
\footnotesize
\caption{Performance variations when different components are applied.}
\label{tab:abl_components}
\begin{tabular}{lccccc}
\hline
\textbf{Method} & \textbf{BLEU-1} & \textbf{BLEU-4} & \textbf{METEOR} & \textbf{ROUGE-L} \\
\hline
Base Model & 0.411 & 0.178 & 0.184 & 0.336 \\
+ NN & 0.428 & 0.193 & 0.194 & 0.354 \\
+ NN + Guideline & 0.429 & 0.195 & 0.195 & 0.355 \\
+ NN + Feedback & 0.429 & 0.193 & 0.195 & 0.354 \\
\textbf{+ NN + Guideline + Feedback} & \textbf{0.431} & \textbf{0.196} & \textbf{0.197} & \textbf{0.357} \\
\hline
\end{tabular}
\end{table}

\paragraph{Ablation Study of Components.}
We evaluate how different components influence performance in Table~\ref{tab:abl_components}. By comparing the first two rows, we see that the evident performance improvement can be obtained when in-context learning is applied, e.g., the BLEU-1 score improved from 0.411 to 0.428, and the ROUGE-L score improved from 0.336 to 0.354. The best results are observed when all components are integrated. 



\begin{figure}
    \centering
    \includegraphics[width=12cm]{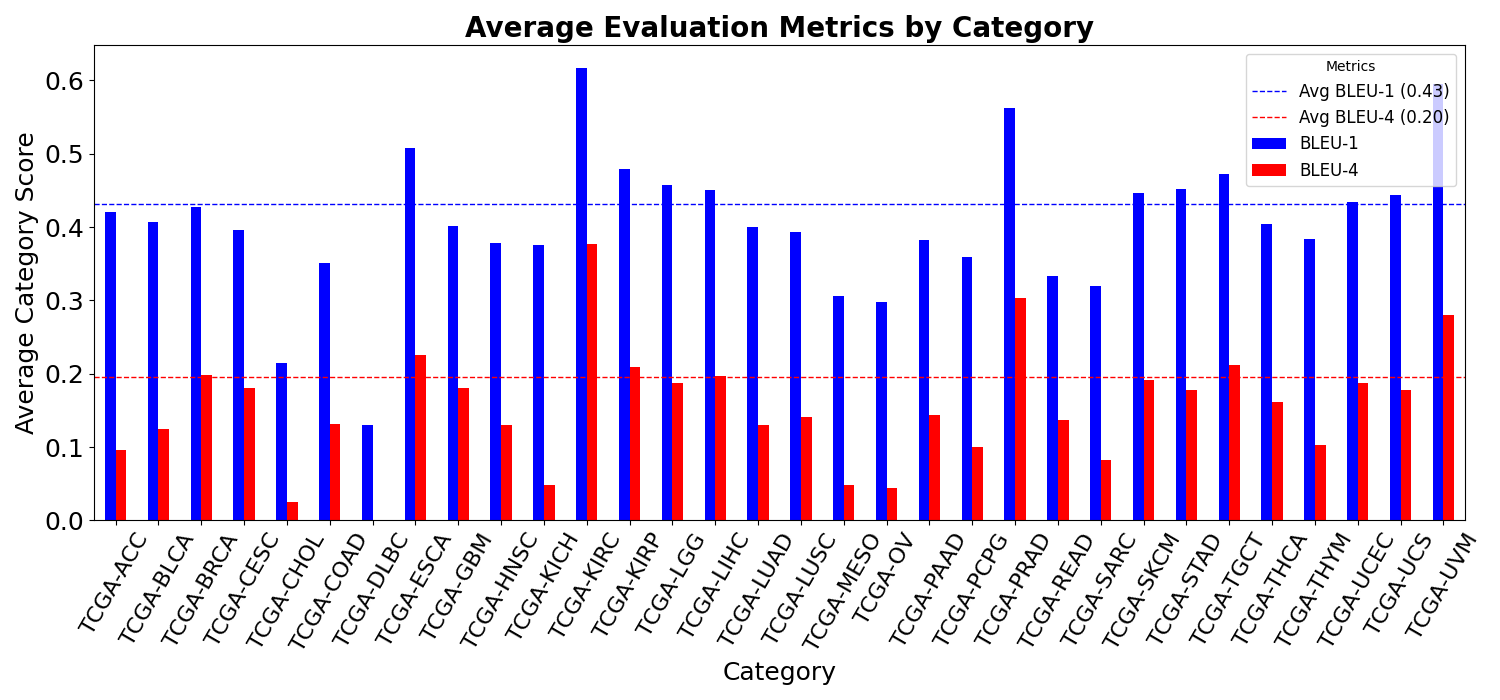}
    \caption{BLEU scores across the 32 disease categories.}
    \label{fig:polygon_diseases}
\end{figure}

\paragraph{Performance on Different Diseases.}
We analyze BLEU scores across 32 disease categories to deepen our understanding of model performance. Figure~\ref{fig:polygon_diseases} presents a bar plot showing BLEU-1 and BLEU-4 scores for different categories, indicating that different diseases pose distinct challenges for our model. 

\subsection{Sample Results}
\begin{figure}
    \centering
    \includegraphics[width=12cm]{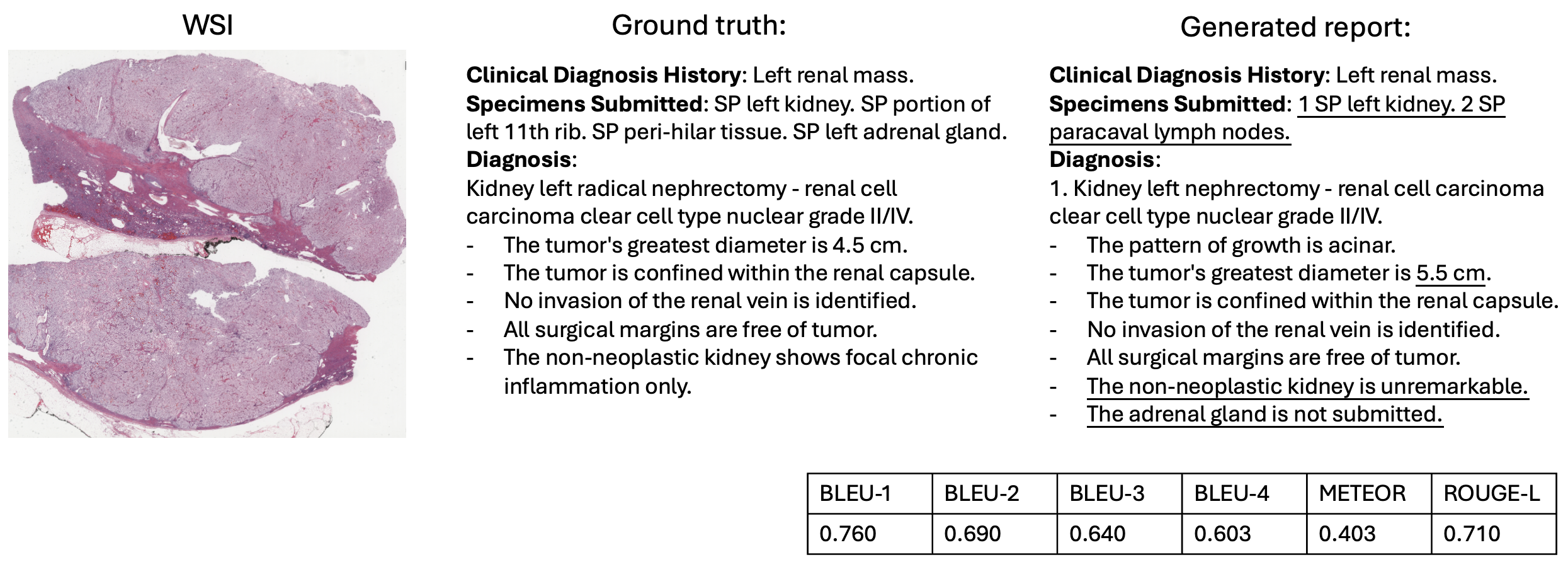}
    \caption{A sample result of the generated report and its corresponding ground truth. The generated result is relatively better.}
    \label{fig:good_sample}
\end{figure}

\begin{figure}
    \centering
    \includegraphics[width=12cm]{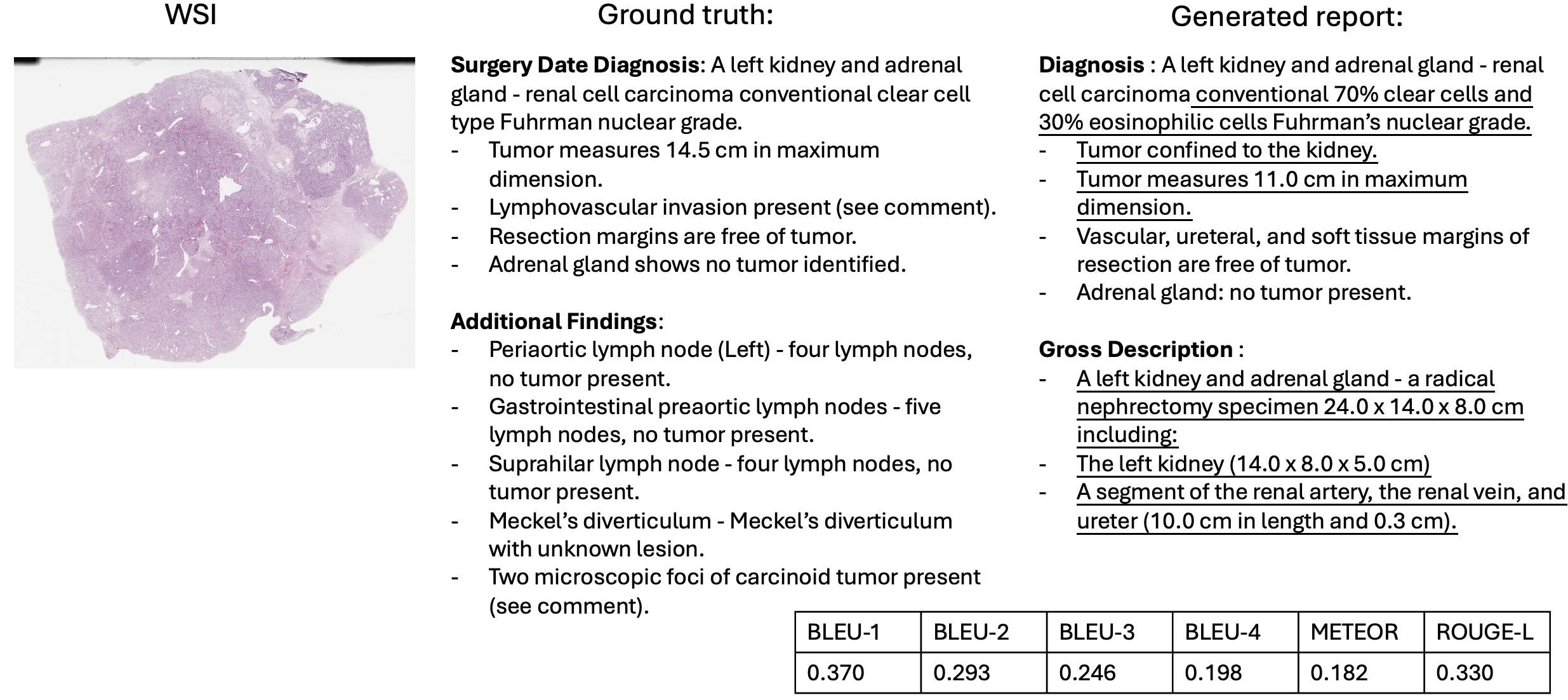}
    \caption{A sample result of the generated report and its corresponding ground truth. The generated result is relatively worse.}
    \label{fig:bad_sample}
\end{figure}

Figure~\ref{fig:good_sample} and Figure~\ref{fig:bad_sample} show sample results of generated reports and their corresponding ground truths, respectively. In both figures, major differences between the ground truth and the generated reports are underlined, and the performance in terms of BLEU, METEOR, and ROUGE-L is shown. Two observations can be made from these two samples. First, although these metrics may not be the most appropriate metrics to evaluate medical report generation, we clearly can see that better generation (Figure~\ref{fig:good_sample}) gives higher values. Second, the main reason for bad results (Figure~\ref{fig:bad_sample}) is not generating incorrect information or irrelevant information but rather missing a few key items in the report.

Both good and bad samples belong to the TCGA-KIRC subset. Based on KNN, both were correctly classified to get TCGA-KIRC guidelines, and the retrieved feedback highlighted several important aspects. However, we found that the main reason for poor performance in Figure~\ref{fig:bad_sample} is the absence of critical medical entities, such as "Periaortic Lymph Nodes". While the nearest-neighbor report helps reduce missing content, it does not fully cover all key items present in the ground truth report. How to more accurately and adaptively guide the generation model with key medical items is thus an important future work. 

\section{Conclusion}
We have introduced \textbf{PathGenIC}, a multimodal in-context learning framework specifically designed for histopathology report generation. This framework generates histopathology reports by a vision language model with multimodal in-context learning. By considering contextual cues in VLMs, this approach significantly enhances VLMs' capacity to generate histopathology reports. We verify the effectiveness of this work by achieving state-of-the-art performance on the HistGen benchmark. In the future, a larger-scale evaluation can be done to demonstrate the generality of the proposed method. More elaborate in-context learning approaches will also be investigated.

\clearpage  
\midlacknowledgments{This work was funded in part by NVIDIA and the National Science and Technology Council, Taiwan, under grants 114-2425-H-006-004, 113-2622-E-006-029, 113-2634-F-006-002, and 112-2221-E-006-136-MY3.}

\bibliography{midl25_049}

\begin{thebibliography}{16}
\providecommand{\natexlab}[1]{#1}
\providecommand{\url}[1]{\texttt{#1}}
\expandafter\ifx\csname urlstyle\endcsname\relax
  \providecommand{\doi}[1]{doi: #1}\else
  \providecommand{\doi}{doi: \begingroup \urlstyle{rm}\Url}\fi

\bibitem[Anderson et~al.(2017)Anderson, He, Buehler, Teney, Johnson, Gould, and
  Zhang]{updown}
Peter Anderson, Xiaodong He, Chris Buehler, Damien Teney, Mark Johnson, Stephen
  Gould, and Lei Zhang.
\newblock Bottom-up and top-down attention for image captioning and visual
  question answering.
\newblock In \emph{Proceedings of IEEE/CVF Conference on Computer Vision and
  Pattern Recognition}, pages 6077--6086, 2017.

\bibitem[Branson et~al.(2021)Branson, Williams, Chan, Graber, Lane, Grieser,
  Landis-Lewis, Cooke, Upadhyay, Mondoux, Singh, Zwaan, Friedman, and
  Olson]{diagnosis_learning_cycle}
Carolina~Fernandez Branson, Michelle Williams, Teresa~M Chan, Mark~L Graber,
  Kathleen~P Lane, Skip Grieser, Zach Landis-Lewis, James Cooke, Divvy~K
  Upadhyay, Shawn Mondoux, Hardeep Singh, Laura Zwaan, Charles Friedman, and
  Andrew P~J Olson.
\newblock Improving diagnostic accuracy through feedback: The diagnosis
  learning cycle.
\newblock \emph{BMJ Quality \& Safety}, 30\penalty0 (12):\penalty0 1002--1007,
  2021.

\bibitem[Brown et~al.(2020)Brown, Mann, Ryder, Subbiah, Kaplan, Dhariwal,
  Neelakantan, Shyam, Sastry, Askell, Agarwal, Herbert-Voss, Krueger, Henighan,
  Child, Ramesh, Ziegler, Wu, Winter, Hesse, Chen, Sigler, Litwin, Gray, Chess,
  Clark, Berner, McCandlish, Radford, Sutskever, and Amodei]{gpt3}
Tom~B. Brown, Benjamin Mann, Nick Ryder, Melanie Subbiah, Jared Kaplan,
  Prafulla Dhariwal, Arvind Neelakantan, Pranav Shyam, Girish Sastry, Amanda
  Askell, Sandhini Agarwal, Ariel Herbert-Voss, Gretchen Krueger, Tom Henighan,
  Rewon Child, Aditya Ramesh, Daniel~M. Ziegler, Jeffrey Wu, Clemens Winter,
  Christopher Hesse, Mark Chen, Eric Sigler, Mateusz Litwin, Scott Gray,
  Benjamin Chess, Jack Clark, Christopher Berner, Sam McCandlish, Alec Radford,
  Ilya Sutskever, and Dario Amodei.
\newblock Language models are few-shot learners.
\newblock In \emph{Proceedings of International Conference on Neural
  Information Processing Systems}, pages 1877--1901, 2020.

\bibitem[Chen et~al.(2024)Chen, Li, Zhu, Zheng, Shui, and Yang]{wsicaption}
Pingyi Chen, Honglin Li, Chenglu Zhu, Sunyi Zheng, Zhongyi Shui, and Lin Yang.
\newblock Wsicaption: Multiple instance generation of pathology reports for
  gigapixel whole-slide images.
\newblock In \emph{Proceedings of International Conference on Medical Image
  Computing and Computer Assisted Intervention}, pages 546--556, 2024.

\bibitem[Chen et~al.(2020)Chen, Song, Chang, and Wan]{r2gen}
Zhihong Chen, Yan Song, Tsung-Hui Chang, and Xiang Wan.
\newblock Generating radiology reports via memory-driven transformer.
\newblock In \emph{Proceedings of Conference on Empirical Methods in Natural
  Language Processing}, pages 1439--1449, 2020.

\bibitem[Chen et~al.(2021)Chen, Shen, Song, and Wan]{r2gencmn}
Zhihong Chen, Yaling Shen, Yan Song, and Xiang Wan.
\newblock Cross-modal memory networks for radiology report generation.
\newblock In \emph{Proceedings of Annual Meeting of the Association for
  Computational Linguistics}, pages 5904--5914, 2021.

\bibitem[Cornia et~al.(2020)Cornia, Stefanini, Baraldi, and Cucchiara]{r2trans}
Marcella Cornia, Matteo Stefanini, Lorenzo Baraldi, and Rita Cucchiara.
\newblock Meshed-memory transformer for image captioning.
\newblock In \emph{Proceedings of the IEEE/CVF Conference on Computer Vision
  and Pattern Recognition}, pages 10578--10587, 2020.

\bibitem[Doe and Smith(2023)]{pathology_peer_review}
John Doe and Alice Smith.
\newblock Anatomic pathology quality assurance through peer review.
\newblock \emph{PathologyOutlines}, 2023.
\newblock https://www.pathologyoutlines.com/topic/managementlabAPQA.html.

\bibitem[Guo et~al.(2024)Guo, Ma, Xu, Wang, Wang, and Chen]{histgen}
Zhengrui Guo, Jiabo Ma, Yingxue Xu, Yihui Wang, Liansheng Wang, and Hao Chen.
\newblock Histgen: A local-global encoding framework for pathology report
  generation.
\newblock In \emph{Proceedings of Medical Image Computing and Computer Assisted
  Intervention}, pages 189--199, 2024.

\bibitem[Hu et~al.(2022)Hu, Shen, Wallis, Allen-Zhu, Li, Wang, Wang, and
  Chen]{lora}
Edward~J Hu, Yelong Shen, Phillip Wallis, Zeyuan Allen-Zhu, Yuanzhi Li, Shean
  Wang, Lu~Wang, and Weizhu Chen.
\newblock Lora: Low-rank adaptation of large language models.
\newblock In \emph{Proceedings of International Conference on Learning
  Representations}, 2022.

\bibitem[Lewis et~al.(2020)Lewis, Perez, Piktus, Petroni, Karpukhin, Goyal,
  Küttler, Lewis, tau Yih, Rocktäschel, Riedel, and Kiela]{rag}
Patrick Lewis, Ethan Perez, Aleksandra Piktus, Fabio Petroni, Vladimir
  Karpukhin, Naman Goyal, Heinrich Küttler, Mike Lewis, Wen tau Yih, Tim
  Rocktäschel, Sebastian Riedel, and Douwe Kiela.
\newblock Retrieval-augmented generation for knowledge-intensive nlp tasks.
\newblock In \emph{Proceedings of Conference on Neural Information Processing
  Systems}, 2020.

\bibitem[Li et~al.(2023)Li, Wong, Zhang, Usuyama, Liu, Yang, Naumann, Poon, and
  Gao]{llavamed}
Chunyuan Li, Cliff Wong, Sheng Zhang, Naoto Usuyama, Haotian Liu, Jianwei Yang,
  Tristan Naumann, Hoifung Poon, and Jianfeng Gao.
\newblock Llava-med: Training a large language-and-vision assistant for
  biomedicine in one day.
\newblock In \emph{Proceedings of Conference on Neural Information Processing
  Systems Track on Datasets and Benchmarks}, 2023.

\bibitem[Miura et~al.(2021)Miura, Zhang, Tsai, Langlotz, and Jurafsky]{miura21}
Yasuhide Miura, Yuhao Zhang, Emily Tsai, Curtis Langlotz, and Dan Jurafsky.
\newblock Improving factual completeness and consistency of image-to-text
  radiology report generation.
\newblock In \emph{Proceedings of the Conference of the North American Chapter
  of the Association for Computational Linguistics: Human Language
  Technologies}, pages 5288--5304, 2021.

\bibitem[Seyfioglu et~al.(2025)Seyfioglu, Ikezogwo, Ghezloo, Krishna, and
  Shapiro]{quilt_llava}
Mehmet~Saygin Seyfioglu, Wisdom~O. Ikezogwo, Fatemeh Ghezloo, Ranjay Krishna,
  and Linda Shapiro.
\newblock Quilt-llava: Visual instruction tuning by extracting localized
  narratives from open-source histopathology videos.
\newblock \emph{arXiv:2312.04746v3}, 2025.

\bibitem[Vaswani et~al.(2017)Vaswani, Shazeer, Parmar, Uszkoreit, Jones, Gomez,
  Kaiser, and Polosukhin]{vaswani17}
Ashish Vaswani, Noam Shazeer, Niki Parmar, Jakob Uszkoreit, Llion Jones,
  Aidan~N. Gomez, Lukasz Kaiser, and Illia Polosukhin.
\newblock Attention is all you need.
\newblock In \emph{Proceedings of Conference on Neural Information Processing
  Systems}, 2017.

\bibitem[Vinyals et~al.(2015)Vinyals, Toshev, Bengio, and Erhan]{showtell}
Oriol Vinyals, Alexander Toshev, Samy Bengio, and Dumitru Erhan.
\newblock Show and tell: A neural image caption generator.
\newblock In \emph{Proceedings of IEEE Conference on Computer Vision and
  Pattern Recognition}, pages 3156--3164, 2015.

\end{thebibliography}

\appendix

\section{Detailed Prompts}
In the feedback scenario, we ask GPT4o as an expert reviewer, provide it with the ground truth and the generated report, and ask it to compare two reports and provide feedback. The detailed prompt is: 

\textsf{You are an expert reviewer specializing in medical report quality assurance. }

\textsf{Ground Truth: \{ground truth\} }

\textsf{Generated Report: \{generated text\}}

\textsf{Comparing the ground truth and generated report, what is the thing that generated report lacks? what suggestion would you give to improve the content of the generated report? The suggestion should be deeply insightful. Be honest and harsh.
}

In the guideline scenario, we ask GPT4o as an AI analyst, provide it the several reports, and ask it to summarize the guidelines to write a report. The detailed prompt is: 

\textsf{You are an advanced AI analyst specializing in deep linguistic and structural analysis of medical reports.}

\textsf{Report 1: \{report\}}

\textsf{Report 2: \{report\}}

\textsf{...}

\textsf{Report 20: \{report\}}

\textsf{Deeply analyze these reports and extract habits, preferences, and especially biases that exist in reports of this category different from general TCGA reports. Your observations must be brutally insightful. Using the insights from observing the habits, preferences, and especially biases in these reports compared with other general TCGA reports. Conclude the habits, preferences, and especially biases with 5 short guidelines that are so insightful even harsh, ensuring anyone reading them knows the exact way to mimic these reports. 
}

\end{document}